\documentclass{article}

\usepackage{arxiv}

\usepackage[utf8]{inputenc} 
\usepackage[T1]{fontenc}    
\usepackage{hyperref}       

\usepackage{url}            
\usepackage{booktabs}       
\usepackage{amsfonts}       
\usepackage{nicefrac}       
\usepackage{microtype}      
\usepackage{lipsum}
\usepackage{graphicx}
\usepackage{multirow}
\usepackage[table,xcdraw]{xcolor}
\graphicspath{ {./images/} }
\usepackage{pifont}
\usepackage{amsmath} 
\usepackage{cleveref} 

\usepackage{algorithmic}
\usepackage{etoolbox} 
\usepackage{subcaption}
\usepackage{algorithm}

\title{Decoupling Semantics from Vision: A Framework for Faithful Visual-Text Compression Evaluation}

\author{
  Yonghan Gao\textsuperscript{1}$^\dagger$,
  Zehong Chen\textsuperscript{1}$^\dagger$,
  Lijian Xu\textsuperscript{1}$^*$,
  Jingzhi Chen\textsuperscript{1},
  Jingwei Guan\textsuperscript{2},
  Xingyu Zeng\textsuperscript{1}$^*$ \\
  \textsuperscript{1}Shenzhen University of Advanced Technology,
  \textsuperscript{2}Shenzhen Technology University \\
  $^\dagger$Equal contribution \\
  $^*$Corresponding authors: \texttt{\{xulijian, zengxingyu\}@suat-sz.edu.cn}
}

\begin{document}

\maketitle

\begin{abstract}
Recent visual-text compression (VTC) methods, typified by DeepSeek-OCR, report impressive high token compression ratios for long-context modeling tasks by leveraging text-to-image rendering. However, existing evaluation protocols heavily rely on downstream task performance. Such evaluation metrics fail to accurately measure text preservation due to the strong inherent linguistic priors of Multimodal Large Language Models (MLLMs). In this work, we introduce a new evaluation framework that decouples MLLMs' capabilities to faithfully assess VTC quality. Within this framework, we further introduce the ZeroSense Benchmark to ensure low semantic correlation of testing samples. By eliminating textual dependencies, our benchmark guarantees that the evaluation results are purely reflective of VTC quality, unaffected by the semantic inference capabilities of downstream models. Extensive experiments across multiple datasets demonstrate that VTC quality and downstream task accuracy diverge significantly, highlighting the necessity of our decoupled evaluation framework.
\end{abstract}


\section{Introduction}

The escalating demand for long-context processing in Large Language Models (LLMs) has encountered a fundamental bottleneck due to the quadratic complexity of self-attention. 
Distinct from traditional compression methods for long-context modeling~\cite{chen-etal-2024-long,chen2024longloraefficientfinetuninglongcontext,chen2025minimax}, the Visual-Text Compression (VTC) paradigm proposes rendering extensive textual sequences into compact document images, effectively substituting thousands of textual tokens with visual tokens. 
Beyond DeepSeek-OCR~\cite{wei2025deepseek}, systems such as Glyph~\cite{cheng2026glyph} and VTC-R1~\cite{wang2026vtcr1visiontextcompressionefficient} have demonstrated significant potential in inference acceleration and context extension.

The rendering process is central to VTC rather than a simple preprocessing step. Font size, spacing, text density, layout, and image resolution jointly determine both the number of visual tokens and the legibility of the rendered content. Different rendering configurations can therefore produce markedly different trade-offs between compression efficiency and textual information preservation. Reliable evaluation of the rendering itself is thus essential for comparing VTC methods and selecting effective configurations under a fixed visual-token budget.
\begin{figure}[t]
  \centering
  \includegraphics[width=0.75\linewidth]{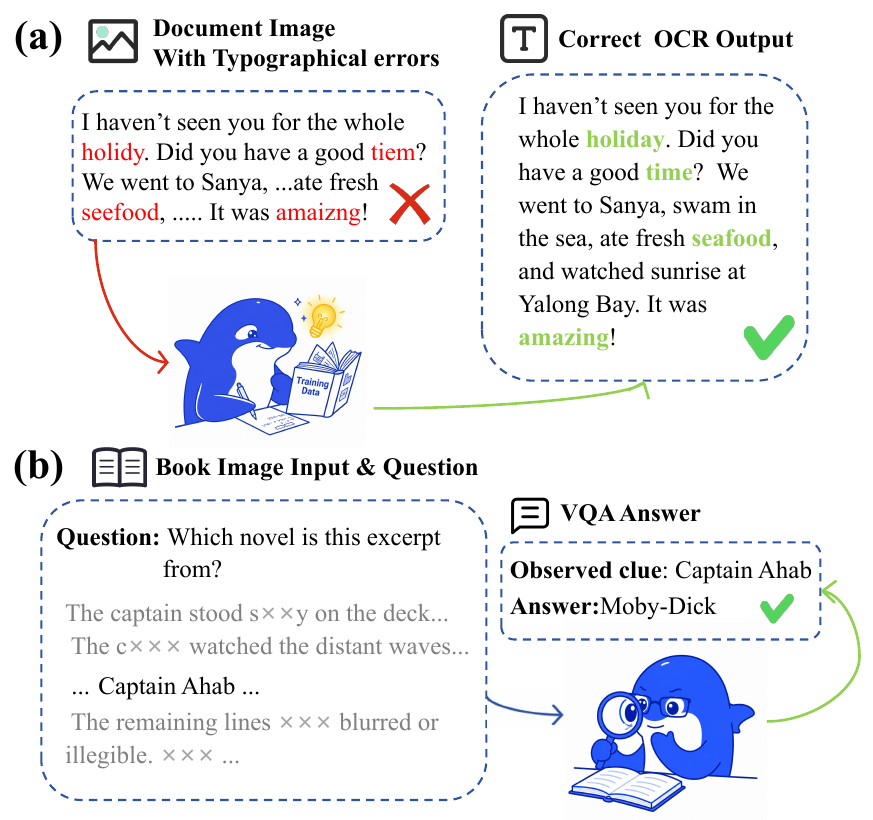}
  
  \caption{Semantic confounds in evaluating visual-text compression. (a) Semantic Priors Compensation: The model leverages contextual inference to rectify ty- pographical errors in the source image. (b) Evidence-Sparse VQA Shortcut: In downstream VQA, a few recognizable keywords can be sufficient to infer the correct answer despite substantial unreadable content, causing task accuracy to overestimate the fidelity of textual information preservation.}
    \label{fig:framework}
    \vspace{-10pt}
\end{figure}
Existing evaluations, however, commonly rely on OCR accuracy over natural text or performance on downstream tasks such as question answering, retrieval, summarization, and reasoning. As illustrated in Fig.~\ref{fig:framework}, these results do not directly measure how much source text remains recoverable from the rendered image. In natural-text OCR, semantic context can help a model correct visually ambiguous content. In downstream tasks such as VQA, a few recognizable clues may be sufficient to infer the answer even when most of the text is unreadable. Moreover, the observed result also depends on the recognition capability of the evaluation model. Consequently, downstream performance reflects the joint effect of rendering quality, semantic inference, task-specific evidence requirements, and evaluator capability, rather than textual preservation alone.

To address these confounding factors, we introduce a unified evaluation framework for VTC. The framework uses complete-text OCR as the evaluation task, constructs source contexts with low semantic predictability, and evaluates each rendering with a fixed set of MLLMs. The final preservation score is obtained from the best model-level result over the complete benchmark, which reduces underestimation caused by the limitations of any single evaluator.

To instantiate semantic-isolated evaluation, we also construct ZeroSense, a novel benchmark characterized by low semantic coherence. By synthesizing text without linguistic structure and rendering it following the visual styles of Fox~\cite{liu2024focusfinegrainedmultipagedocument}, Omni~\cite{Ouyang_2025_CVPR} and Glyph, ZeroSense allows the model's performance to approximate the pure effect of visual fidelity. 

Finally, we apply the proposed framework to evaluate three VTC rendering functions across multiple compression ratios. Surprisingly, the text-preservation estimates produced by our framework differ substantially from the conventional results reported by DeepSeek-OCR. For example, under the FOX layout at a \(10\times\) compression ratio, DeepSeek-OCR achieves an end-to-end OCR accuracy of 95.4\%, whereas our framework estimates a text-preservation rate of only 11.8\%. This discrepancy demonstrates that high end-to-end OCR accuracy does not necessarily indicate that the visual compression process has preserved a comparable amount of source-text information, highlighting the need for a dedicated evaluation framework for VTC text preservation.

In summary, our contributions are threefold:

\begin{itemize}
    \item We propose a framework for evaluating text preservation in VTC. Its estimates differ substantially from prior end-to-end results and better reflect textual preservation capability.

    \item We introduce ZeroSense, a dataset of long textual contexts with low textual predictability. Its source texts are independent of any image representation, enabling the same content to be evaluated across different VTC rendering functions and compression settings.

    \item We apply the proposed framework to several VTC methods, revealing substantial differences in their text-preservation capabilities and performance degradation as compression increases.
\end{itemize}

\section{Related Work}

\subsection{Visual-Text Compression Methods}
\label{sec:related_vtc}

Before VTC, several studies had already explored processing text directly from images. Pixel-based language models read rendered text from pixels. OCR-free document models directly predicted text or task outputs from document images~\cite{kim2022ocr,rust2023language,lee2023pix2struct,kesen2025multilingual}. These studies showed that visual representations can carry textual information. However, they were not designed to compress long textual contexts.
Later studies used visual representations to reduce the cost of long-context processing Before VTC, several studies had already explored processing text directly from images \cite{Xing2025SeeTT}. They rendered text as images and encoded the images into visual tokens. Early methods showed that visual tokens can extend the context available to multimodal models~\cite{wang2024leveraging,lu2024textpixeladvancinglongcontext}. Several methods combined text tokens with visual tokens or selected which parts of the context should be visually encoded~\cite{xing2026vision,li-etal-2025-text}. 

Recent systems have developed this idea into dedicated VTC methods. DeepSeek-OCR studies high-ratio optical compression through document reconstruction, while Glyph combines an optimized rendering procedure with a vision-language model trained for general long-context tasks. Subsequent work has extended VTC to global context modeling, reasoning, and selective context expansion~\cite{jiao2026globalcontextcompressioninterleaved,wang-etal-2026-render,xie2026lensvlmselectivecontextexpansion,gao2026zerosense}. These studies establish VTC as a general approach to long-context processing, with different rendering procedures, models, and evaluation tasks, while related efforts have also explored semantic-aware visual representation learning through semantic alignment and adaptive token selection strategies~\cite{yang2025one,young2026scalar,he2026beyond,he2026autoselect,he2026stepwise,young2026fewer}.

\subsection{Evaluation of Visual-Text Compression}
\label{sec:related_evaluation}

Existing VTC evaluations mainly use many downstream tasks. Text recovery measures whether a model can reproduce text from a compressed document image. It is usually evaluated on natural document pages or document benchmarks. Other downstream evaluation measures whether the visual context supports tasks such as retrieval, question answering, reasoning, and memory~\cite{hsieh2024rulerwhatsrealcontext,bai2023longbench,zhao2025vtcbenchvisionlanguagemodelsunderstand}.

Concurrent studies have examined VTC results beyond reported task scores, including semantic dependence, encoder design, and task-specific information loss~\cite{tu-etal-2026-fico,liang2026visualmeritlinguisticcrutch,lee2026opticalcontextcompressionjust,tang2026visualtextcompressionmeasure}. However, they do not provide a controlled quantitative measure of how much information is preserved by a VTC rendering function and our work fills this gap.
\begin{figure*}
    \centering
    \includegraphics[width=0.95\linewidth]{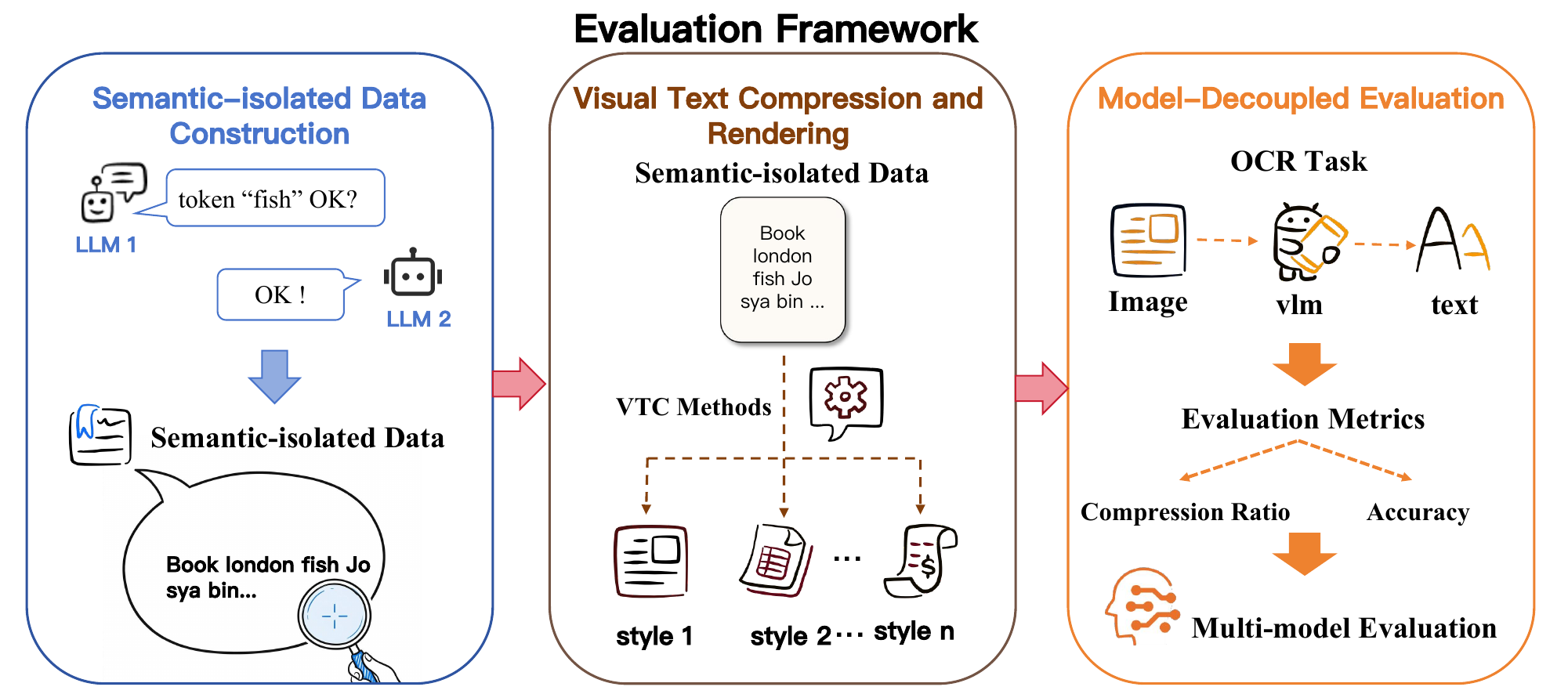}
    \caption{This overview illustrates our proposed evaluation framework. ZeroSense constructs low-predictability texts via probability-constrained generation and multi-model validation to reduce semantic compensation. The generated texts are further processed with diverse VTC rendering strategies and compression configurations. Finally, multiple MLLMs perform OCR-based text reconstruction, and the maximum recovery accuracy is adopted to quantify the context preservation ability of each method.
}
    \label{fig:zerosense_framework}
\end{figure*}

\section{Evaluation Framework}
\subsection{Problem Formulation}
\label{sec:problem_formulation}

A good VTC method aims to transform a long textual context into a compact visual representation while preserving the context required for downstream tasks. Formally, let a dataset be defined as 
\begin{equation}
D=\left\{(q_i,c_i)\right\}_{i=1}^{N},
\end{equation}
where $q_i$ denotes a query (or instruction) and $c_i$ denotes the corresponding long textual context required to answer $q_i$. A rendering function $R(\cdot)$ converts the textual context into a visual representation,
\begin{equation}
\label{2}
I_{R,i}=R(c_i),
\end{equation}
Given an arbitrary multimodal large language model M, let $f_M(\cdot, \cdot)$ denote its inference process. The primary objective of VTC is to ensure that replacing the original textual context with its rendered visual representation does not significantly alter the model's behavior, i.e.,
\begin{equation}
f_M(q_i,c_i)\approx f_M(q_i,I_{R,i}),
\qquad
\forall\,M,\ \forall\,(q_i,c_i)\in D.
\end{equation}
where the approximation can be measured by prediction consistency or downstream task performance.

More generally, let T denote an arbitrary downstream task, and let $\mathcal{P}_T(M,X)$ represent the performance of model M on task T using context representation X. The ideal VTC objective is to preserve task performance,
\begin{equation}
\mathcal{P}_T(M,c_i)
\approx
\mathcal{P}_T(M,I_{R,i}),
\qquad
\forall\,T,\ \forall\,M.
\end{equation}

Specifically, an effective VTC method should maintain comparable performance across arbitrary datasets and downstream tasks while minimizing the number of visual tokens required to represent the context. Let  $N_{\mathrm{text}}(\cdot)$ and $N_{\mathrm{vis}}(\cdot)$ denote the numbers of text tokens and visual tokens consumed by the model, respectively. The compression objective can be formulated as:

\begin{equation}
\begin{aligned}
\label{5}
R^{*}
=
\arg\max_{R}\quad &
\mathbb{E}_{(q,c)\sim D}
\left[
\frac{N_{\mathrm{text}}(c)}
     {N_{\mathrm{vis}}(R(c))}
\right] \\
\text{s.t.}\quad &
\mathcal{P}_T(M,R(c))
\approx
\mathcal{P}_T(M,c),
\quad
\forall\,M,\ \forall\,T.
\end{aligned}
\end{equation}

Therefore, the fundamental goal of Visual Text Compression is to learn a rendering method that achieves the highest possible compression ratio while remaining context-preserving for arbitrary downstream tasks and models.

\subsection{Evaluation Requirements}
\label{subsec:evaluation_requirements}

According to the formulation above, evaluating a VTC method requires jointly assessing two orthogonal objectives: compression efficiency and context preservation. While compression efficiency can be directly quantified by the visual-text token compression ratio in Eq.~(\ref{5}), it is more challenging to faithfully measure context preservation. The ideal objective requires preserving textual information across arbitrary downstream tasks and arbitrary multimodal language models, which is infeasible in practice. Therefore, an effective evaluation approach should approximate this objective while minimizing the influence of evaluation bias. Based on the formulation above, we argue that a reliable VTC evaluation should satisfy the following three principles.

\paragraph{Fundamental Downstream Task Selection}
Since it is impossible to evaluate every downstream task, the selected evaluation tasks should directly measure how much the textual information is preserved, rather than the reasoning capability of a particular model. Most text-centric downstream tasks, such as question answering, information retrieval, summarization, and reasoning, fundamentally rely on correctly perceiving the textual content before higher-level reasoning can be performed. Therefore, textual recognition constitutes a necessary prerequisite for these tasks. Motivated by this observation, we adopt optical character recognition as the evaluation task and the OCR performance as the primary measure of recoverable context preservation.
\paragraph{Model-Decoupled Evaluation}
According to Eq.~(\ref{5}), the objective of VTC is independent of any specific model. However, different MLLMs exhibit substantially different OCR capabilities \cite{Liu2023OCRBenchOT}. It would inevitably entangle rendering quality with model capability if we directly compare OCR accuracy obtained by a single model or averaging across multiple models. Since our objective is to evaluate the recoverable context preserved by the rendering itself rather than the recognition ability of a particular model, we estimate textual preservation using the maximum OCR performance achieved across a diverse set of MLLMs. Intuitively, if the rendered image faithfully preserves the textual information, at least one sufficiently capable model should be able to recover it. Taking the maximum performance therefore provides a better approximation to the upper bound of recoverable textual information while reducing the confounding effect introduced by model capability. 
\paragraph{Semantic-Isolated Evaluation Data}
Modern MLLMs possess strong prior knowledge and reasoning \cite{pmlr-v235-lin24c}. Even when part of the textual information is lost during compression, models may still infer the missing content from semantic context, leading to overestimation of textual preservation. Such semantic compensation reflects reasoning ability rather than textual preservation, making it unsuitable for evaluating VTC. Therefore, the evaluation data should minimize semantic dependency among text elements so that each text unit can only be recovered from the rendered image itself, rather than inferred from textual knowledge.

Based on these three principles, we propose an evaluation framework introduced in Section \ref{sec:framework_design} that jointly evaluates compression efficiency and textual preservation while minimizing the confounding effects introduced by downstream task selection, model capability, and semantic reasoning.

\subsection{Framework Design}
\label{sec:framework_design}
Guided by the principles above, we design a unified evaluation framework that jointly measures compression efficiency and textual preservation. As illustrated in Fig.\ref{fig:zerosense_framework}, the proposed framework consists of three sequential stages: semantic-isolated data construction, visual text compression and rendering, and model-decoupled evaluation.

\textbf{Stage1: Semantic-isolated Data Construction}
Unlike natural documents where textual elements are often semantically correlated, we need a dataset that minimizes semantic dependency among text units so that each character or word can only be recovered from the visual evidence presented in the rendered image. The detailed construction procedure is described in the next Section~\ref{sec:benchmark_construction}.

\textbf{Stage2: Visual Text Compression and Rendering}
For each evaluated VTC method R, the source context \(c_i\) is processed by \(R\) as defined in (\ref{2}). The resulting representation is then passed to the subsequent evaluation stage.




\textbf{Stage 3: Model-Decoupled Evaluation}
The evaluation stage consists of two steps: OCR inference and metric computation.
Let \(\mathcal{M}\) denote the fixed set of evaluation MLLMs.
Each \(M\in\mathcal{M}\) independently evaluates all \(N\) source contexts. 
For each rendered image \(I_{R,i}\), we perform OCR using a fixed instruction \(q_{\mathrm{OCR}}\) that requests the model to transcribe all visible text. The recognized text produced by model \(M\) is
\begin{equation}
\hat{c}_{i,M,R}
=
f_M\!\left(q_{\mathrm{OCR}},I_{R,i}\right).
\end{equation}
The recognized text \(\hat{c}_{i,M,R}\) is then compared with the corresponding source context \(c_i\) to compute the evaluation metrics. 

Compression efficiency is directly measured by the visual-text token compression ratio defined in Eq.~(\ref{5}). Textual information preservation is quantified by the average recovery character accuracy over the dataset:
\begin{equation}
\label{7}
A_M(R)
=
\frac{1}{N}
\sum_{i=1}^{N}
\operatorname{Acc}
\left(
\hat{c}_{i,M,R},
c_i
\right),
\end{equation}

where \(\operatorname{Acc}\) denotes the text recovery character accuracy between the recognized text and the source context.
Following the model-decoupled evaluation principle, the final recovery score of method, the final score of \(R\) is
\begin{equation}
A(R)
=
\max_{M\in\mathcal{M}} A_M(R).
\end{equation}

Finally, the framework reports the visual-text token compression ratio together with the recovery accuracy \(A(R)\). These two metrics jointly characterize the compression–preservation trade-off of each VTC method.

\begin{figure}[t]
    \centering
    \makebox[\linewidth][c]{\includegraphics[width=0.8\linewidth]{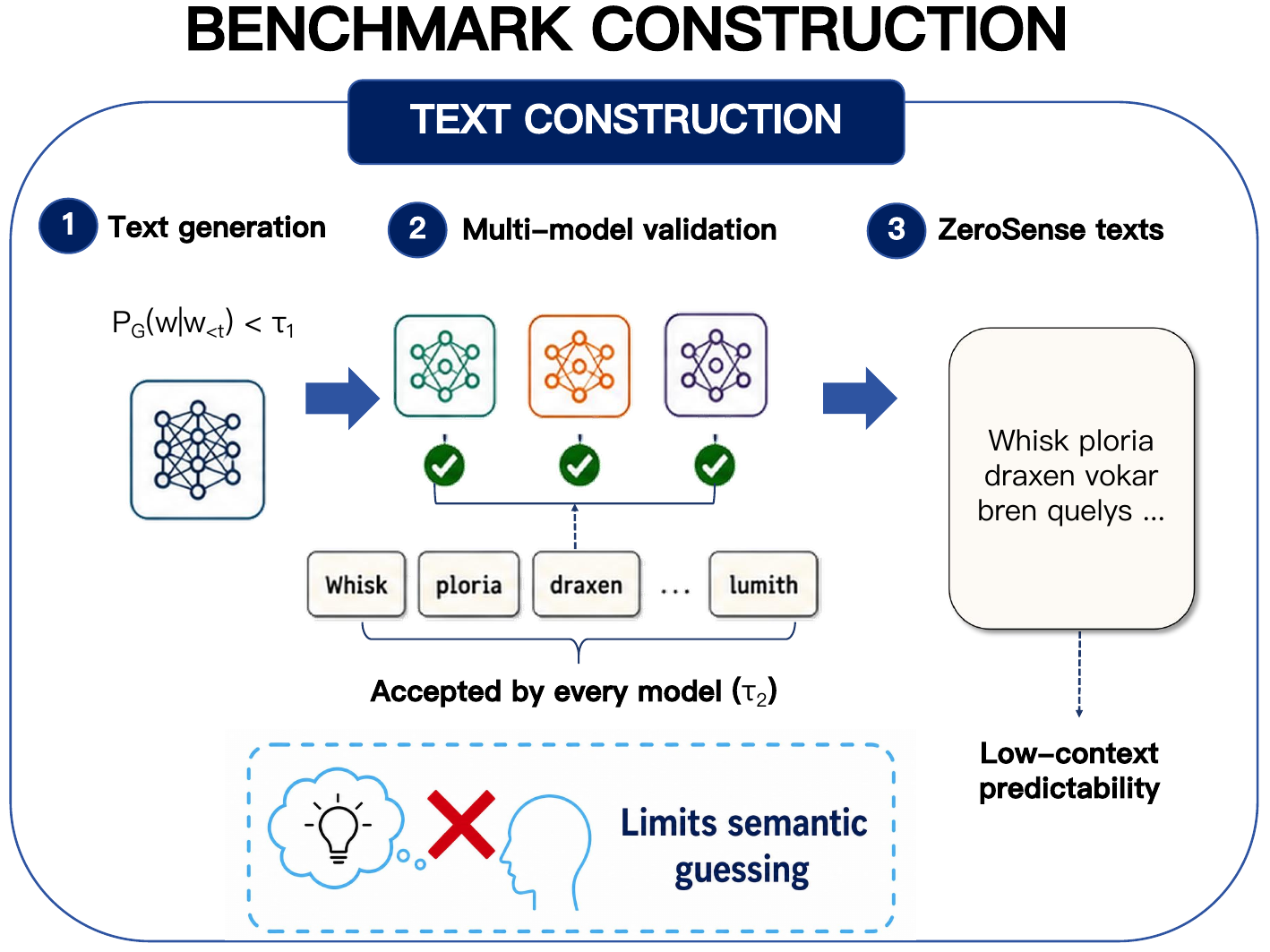}}
    \caption{ZeroSense text construction pipeline. Low-probability words are sampled to form long sequences, which are retained only after multi-model validation, yielding contexts with low textual predictability.}
    \label{fig:zerosense_benchmark}
\end{figure}

\section{ZeroSense Benchmark}
\label{sec:zerosense}

ZeroSense is a text-centered benchmark for evaluating text preservation in VTC. It consists of source contexts with low contextual predictability, which reduces the influence of semantic priors during evaluation. In this section, we will introduce text construction process and benchmark statistics.

\subsection{Benchmark Construction}
\label{sec:benchmark_construction}

\paragraph{Text Construction}
\label{sec:text_construction}

ZeroSense source texts are constructed through probability-constrained generation and multi-model validation. The construction pipeline is depicted in Fig.\ref{fig:zerosense_benchmark}.

Let \(V_{\mathrm{valid}}\) be the intersection of the vocabularies of all used models. A word is included only if it is represented as a single token by every corresponding tokenizer \cite{chai-etal-2024-tokenization}. This provides a common unit for calculating conditional probabilities across models. 
Then we use the causal language model component of GOT-OCR2~\cite{wei2024general} as the generation model \(G\). At step \(t\), the candidate set is obtained with a threshold $\tau_1$.

\begin{equation}
Q_t
=
\left\{
w\in V_{\mathrm{valid}}
\mid
P_G(w\mid w_{<t})<\tau_1
\right\}.
\end{equation}

The next word is sampled uniformly from \(Q_t\). This process produces a sequence \(W=(w_1,\ldots,w_n)\). We generate sequences with approximate target lengths of 300, 500, 700, and 900 words.
A sequence with low probability under \(G\) may remain predictable to other models. We therefore validate each sequence using the causal language model components of Glyph, DeepSeek-OCR, and Hunyuan. Let \(\mathcal{L}\) denote the set of validation models. A sequence is retained only when the averaged probability is less than a threshold $\tau_2$:

\begin{equation}
\frac{1}{n}
\sum_{t=1}^{n}
P_{L_k}(w_t\mid w_{<t})
<
\tau_2,
\qquad
\forall L_k\in\mathcal{L}.
\end{equation}

Each retained sequence forms a source context \(c_i\).



\subsection{Benchmark Statistics}
\label{sec:benchmark_statistics}

ZeroSense contains 12,324 source contexts. \Cref{tab:statistics} reports their distribution according to the approximate text length.

\begin{table}[t]
\centering
\small
\renewcommand{\arraystretch}{1.2} 
\setlength{\tabcolsep}{16pt}      

\begin{tabular}{l r}
\toprule
\textbf{Texts Length} & \textbf{Contexts Count} \\
\midrule
300 words & 4,493 \\
500 words & 3,625 \\
700 words & 2,733 \\
900 words & 1,473 \\
\midrule
\textbf{Total} & \textbf{12,324} \\
\bottomrule
\end{tabular}
\caption{Distribution of source contexts in ZeroSense.}
\label{tab:statistics}
\end{table}

To facilitate direct evaluation, we additionally provide rendered VTC instances based on the FOX layout, the OmniDocBench layout, and the Glyph rendering procedure. Each rendering method is applied to all 12,324 source contexts, resulting in 12,324 images per method and 36,972 images in total. Every rendered image $(I_{R,i})$ is paired with its source context $(c_i)$, which serves as the target for OCR evaluation. The three rendered sets contain identical source content and differ only in their visual representation.

\section{Experiment}
\subsection{Set up}
\paragraph{Text sources.}We use ZeroSense as the core corpus for evaluating context preservation, with coherent natural texts from FOX and OmniDocBench as control groups. Distinct from these readable benchmark texts, ZeroSense adopts low-predictability content. We report results for all three datasets independently, but only ZeroSense measurements are used to compute our context preservation metric. In our implementation, $\tau_1 = 1\times 10^{-8}$ and $\tau_2 = 1\times 10^{-5}$. 

\paragraph{Rendering methods.}We consider three rendering methods: the document layout derived from FOX, the document layout derived from OmniDocBench, and the rendering method introduced by Glyph. Each method is applied to ZeroSense, FOX, and OmniDocBench, producing nine combinations of text source and rendering method.

\paragraph{Evaluation models.} We use DeepSeek-OCR, Qwen2.5-VL~\cite{Qwen2.5-VL}, and HunyuanOCR~\cite{team2025hunyuanocr} as the evaluation model set $\mathcal{M}=\{\text{DeepSeek-OCR},\text{Qwen2.5-VL},\text{HunyuanOCR}\}$. Each model receives a visual input $I_{R,i}$ and recovers its complete textual content as $\widehat{c}_{i,R,M}$. No model is trained or adapted using ZeroSense. All inference experiments are conducted on two NVIDIA RTX A6000 GPUs. We use the same recovery instruction whenever supported by the model interface. Model versions, inference parameters, and prompts are reported in the appendix.

\paragraph{Metric.}
We instantiate \(\operatorname{Acc}\) in
Eq. ~\ref{7} as word-level OCR
accuracy between the recovered text and the source context.  
Similarly, we adopt the formula \(\frac{N_{\mathrm{text}}(c)}{N_{\mathrm{vis}}(R(c))}\) defined in Equation~\ref{5} to compute the compression ratio metric.

\subsection{Main Results}
\label{sec:main_results}
 We apply the proposed evaluation pipeline to three rendering methods: the FOX layout, the OmniDocBench layout, and Glyph method. For each setting, we evaluate complete text recovery using both natural document text and ZeroSense. We then aggregate the results across the fixed model set according to our framework. This experiment compares the resulting estimates of context preservation with those obtained through the conventional evaluation on natural text.

\begin{table*}[t]
\centering

\small
\setlength{\tabcolsep}{8pt}
\begin{tabular}{ll ccc ccc}
\toprule
& &
\multicolumn{3}{c}{\textbf{Ours}} &
\multicolumn{3}{c}{\textbf{DeepSeek-OCR}} \\
\cmidrule(lr){3-5} \cmidrule(lr){6-8}
Rendering Setting & Compression
& FOX & Omni & ZeroSense
& FOX & Omni & ZeroSense \\
\midrule

\multirow{6}{*}{FOX layout}
& \(5\times\)    & 97.7 & 75.2 & 33.8 & 97.7 & 75.2 & 33.8 \\
& \(7.5\times\)  & 96.6 & 70.1 & 20.2 & 96.6 & 70.1 & 20.2 \\
& \(10\times\)   & 95.4 & 65.3 & 11.8 & 95.4 & 65.3 & 11.8 \\
& \(12.5\times\) & 93.8 & 61.2 & 7.0 & 93.8 & 61.2 & 7.0 \\
& \(15\times\)   & 91.8 & 56.6 & 4.8 & 91.8 & 56.6 & 4.8 \\
& \(17.5\times\) & 87.0 & 51.4 & 2.1 & 87.0 & 51.4 & 2.1 \\

\midrule

\multirow{6}{*}{OmniDocBench layout}
& \(5\times\)    & 90.7 & 76.7 & \textcolor{red}{42.8} & 90.7 & 76.7 & 29.1 \\
& \(7.5\times\)  & 82.7 & 76.0 & 17.1 & 82.7 & 76.0 & 17.1 \\
& \(10\times\)   & 74.1 & \textcolor{red}{78.3} & 12.2 & 74.1 & 75.6 & 12.2 \\
& \(12.5\times\) & 68.6 & \textcolor{red}{78.7} & 11.3 & 68.6 & 75.6 & 11.3 \\
& \(15\times\)   & 61.5 & \textcolor{red}{80.5} & 11.3 & 61.5 & 61.5 & 11.3 \\
& \(17.5\times\) & 57.3 & \textcolor{red}{84.2} & 11.2 & 57.3 & 57.3 & 11.2 \\

\midrule

\multirow{6}{*}{Glyph rendering}
& \(5\times\)    & 98.9 & 94.2 & 65.3  & 98.9 & 94.2 & 65.3\\
& \(7.5\times\)  & 98.2 & 92.3 & 52.0 & 98.2 & 92.3 & 52.0\\
& \(10\times\)   & 97.4 & 90.2 & 39.6 & 97.4 & 90.2 & 39.6 \\
& \(12.5\times\) & 96.7 & 87.9 & 29.7  & 96.7 & 87.9 & 29.7 \\
& \(15\times\)   & 94.4 & 86.3 & 23.5 & 94.4 & 86.3 & 23.5 \\
& \(17.5\times\) & 92.3 & 84.0 & 15.1 & 92.3 & 84.0 & 15.1 \\

\bottomrule
\end{tabular}
\caption{
Word-level accuracy across different rendering methods and compression ratios.
FOX and OmniDocBench are natural-text controls, while ZeroSense is used for context preservation evaluation. Numbers marked in red denote results obtained by our proposed method that differ substantially from those yielded by DeepSeek.}
\label{tab:overall_results}
\end{table*}

\subsubsection{The impact of semantic prior}

\Cref{tab:overall_results} reports the results for three types of source context. The accuracy on ZeroSense is markedly lower than that on the source contexts from FOX and OmniDocBench under any rendering method and its corresponding different compression ratios.
The results obtained on FOX source contexts with the FOX layout and on OmniDocBench source contexts with the OmniDocBench layout, defined as prior evaluation paradigm, can represent the  context preservation in the previous report. Across all three evaluation models, these prior evaluation paradigms achieve strong performance. In particular, Glyph maintains a context-preservation score of 92.3\% on the DeepSeek-OCR model even at a compression ratio of $17.5\times$. Surprisingly, when evaluated on ZeroSense, the corresponding score drops sharply to 15.1\%. Moreover, each method's performance declines substantially even at compression ratios previously considered low. For example, the score of FOX on the DeepSeek-OCR model at a compression ratio of 17.5 decreases from nearly 87\% to 7\%.
These results show that the source text has a substantial effect on the observed recovery accuracy. Natural document text contains textual regularities that may help a model recover visually uncertain content. ZeroSense limits this contribution by using text with low textual predictability. The lower ZeroSense scores are therefore consistent with reduced semantic compensation during text recovery. Evaluation on natural text alone may consequently provide a more optimistic estimate of contexts preservation. Section~\ref{sec:benchmark_construction} examines the effect of text construction under more closely controlled conditions. We present the varying trends of experimental results under different thresholds (i.e., different degrees of semantic relevance) in the ablation study~\ref{sec:ablation}.

\subsubsection{Context preservation of different methods}
With the framework, we can obtain more reliable text preservation for different VTC methods. As shown in \Cref{tab:overall_results}, the Glyph rendering method achieves stronger context preservation capability than the other two alternatives from the compression of 5 to 17.5. This demonstrates that the Glyph-based rendering method achieves better context preservation performance than the other two methods. 

We also observe that the FOX layout only outperforms the OmniDocBench layout under the compression ratios of $7.5\times$.
However, in the old evaluation paradigm, the FOX layout would consistently outperform the OmniDocBench layout. This validates the necessity of our evaluation framework.

It is worth noting that the metric value under the OmniDocBench layout at the compression ratio of $5\times$ is higher than those at other compression ratios. This abnormal peak value is obtained by taking the maximum result over multiple evaluation models, which specifically reflects the performance of the Qwen2.5-VL model under $5\times$ compression. The reason why such a high score cannot be obtained at higher compression ratios stems from the intrinsic configuration limitations of Qwen2.5-VL: the model fails to reach higher compression ratios under the fixed text length setting.

Although Qwen2.5-VL only supports relatively low compression ratios, it achieves favorable performance across all tested rendering methods. Detailed numerical data are summarized and presented in the Appendix. These phenomena collectively validate the necessity of taking the maximum score across multiple evaluation models.



\subsection{Ablation Studies}
\label{sec:ablation}

Here we examine two questions. First, we investigate the stability of the evaluation results across different evaluation set sizes. 
Second, we examine whether the main components of the ZeroSense construction process are necessary for reducing contextual predictability. Unless otherwise specified, the rendering methods, compression settings, and evaluation models are kept consistent with the main experiments.

\subsubsection{Stability of the Evaluation Results}

First, we conduct an investigation into the impact of dataset size on experimental performance. Specifically, we break down this factor into two dimensions: context length and benchmark scale. Through these experiments, we aim to explore the stable conditions required for robust evaluation, as well as the variation trends correlated with dataset scale. 

\paragraph{Effect of context length}


We explore how text length affects the final conclusions.
As shown in \Cref{tab:glyph_cr_token_length}, the Glyph performance degrades as the text length increases when the compression ratio remains fixed. Under the ablation setup with text lengths ranging from 200 to 1000 tokens, taking the Glyph rendering method as an example, its capability shows a declining trend with increasing text length under the same compression ratio. When the text length is approximately between 200 and 600 tokens, this rendering method maintains above-average performance. In contrast, lengths exceeding 600 tokens impose adverse effects across all experimental groups. The complete table showing the performance of different rendering methods is provided in the Appendix.

\begin{table}
\centering
\small
\setlength{\tabcolsep}{6pt}
\begin{tabular}{c cccc}
\toprule
\textbf{Compression} & \multicolumn{4}{c}{\textbf{Text Token Length}} \\
\cmidrule(lr){2-5}
\textbf{Ratio} & \textbf{200--400} & \textbf{400--600} & \textbf{600--800} & \textbf{800--1000} \\
\midrule
$2.5\times$  & 71.6 & 68.2 & 64.4 & 59.1 \\
$5.0\times$  & 67.8 & 65.8 & 63.6 & 57.6 \\
$7.5\times$  & 54.8 & 51.7 & 50.0 & 48.2 \\
$10.0\times$ & 42.7 & 39.2 & 37.5 & 36.2 \\
$12.5\times$ & 32.1 & 29.1 & 27.3 & 27.4 \\
$15.0\times$ & 27.0 & 25.3 & 23.3 & 21.1 \\
\bottomrule
\end{tabular}
\caption{ZeroSense precision of the Glyph layout under different compression ratios and text token lengths.}
\label{tab:glyph_cr_token_length}
\par\smallskip
\end{table}

\paragraph{Effect of benchmark size}

Following the investigation on individual context length, we further explore whether the benchmark dataset has a sufficient number of samples to yield stable and reliable evaluation results. As \Cref{tab:glyph_convergence} shows, we randomly sample subsets with different context quantities from the ZeroSense dataset, with sample sizes set to 1000, 4000, 7000 and 10000 respectively.

Taking the Glyph method as an example again, under compression ratios ranging from $2.5\times$ to $15\times$ and with identical text length distributions controlled, the performance maintains minor fluctuations with a deviation no greater than 0.2\% when the sample size increases from 1000 to 10000. All other rendering methods also exhibit stable performance across this sample size range. This demonstrates that the sample volume of the ZeroSense benchmark is sufficient to guarantee stable evaluation outcomes.
\begin{table}
\centering
\small
\setlength{\tabcolsep}{6pt}
\begin{tabular}{c cccc}
\toprule
\textbf{Compression} & \multicolumn{4}{c}{\textbf{Sample Size ($N$)}} \\
\cmidrule(lr){2-5}
\textbf{Ratio} & \textbf{1,000} & \textbf{4,000} & \textbf{7,000} & \textbf{10,000} \\
\midrule
$2.5\times$  & 0.6742 & 0.6761 & 0.6762 & 0.6750 \\
$5.0\times$  & 0.6496 & 0.6487 & 0.6494 & 0.6489 \\
$7.5\times$  & 0.5204 & 0.5195 & 0.5201 & 0.5206 \\
$10.0\times$ & 0.3962 & 0.3951 & 0.3952 & 0.3948 \\
$12.5\times$ & 0.2968 & 0.2959 & 0.2966 & 0.2970 \\
$15.0\times$ & 0.2339 & 0.2346 & 0.2343 & --     \\
\bottomrule
\end{tabular}
\caption{Glyph max precision convergence across sample sizes and compression ratios.}
\label{tab:glyph_convergence}
\end{table}

\subsubsection{Effectiveness of ZeroSense Construction}

The previous experiments examine the stability of the evaluation results. We next study whether the ZeroSense construction procedure effectively reduces contextual predictability. We first compare probability-constrained generation with text shuffling, then analyze the strength of the probability constraint, and finally examine the role of multi-model validation.

\paragraph{Comparison of text construction strategies.}
\begin{table}[t]
\centering

\small
\setlength{\tabcolsep}{4pt} 
\begin{tabular}{c cccc}
\toprule
\textbf{Compression} & \multicolumn{4}{c}{\textbf{Method}} \\
\cmidrule(lr){2-5}
\textbf{Ratio} & \textbf{Swap5} & \textbf{Swap10} & \textbf{Shuffle5} & \textbf{Shuffle10} \\
\midrule
$2.5\times$  & 0.937 & 0.915 & 0.935 & 0.917 \\
$5.0\times$  & 0.849 & 0.819 & 0.842 & 0.812 \\
$7.5\times$  & 0.824 & 0.806 & 0.827 & 0.798 \\
$10.0\times$ & 0.772 & 0.742 & 0.763 & 0.750 \\
$12.5\times$ & 0.730 & 0.744 & 0.734 & 0.750 \\
$15.0\times$ & 0.710 & 0.667 & 0.703 & 0.646 \\
\bottomrule
\end{tabular}
\caption{Experimental results of different text shuffling strategies under varying compression ratios (Glyph layout). \emph{Swap$k$} randomly swaps pairs of adjacent words up to distance $k$, while \emph{Shuffle$k$} performs local random permutation within sliding windows of size $k$.}
\label{tab:shuffle_cr_result}
\end{table}
As \Cref{tab:shuffle_cr_result} and \Cref{tab:zeroprior_threshold_cr} show, a simple way to reduce semantic coherence is to shuffle units from natural text. However, shuffling does not explicitly control the conditional probability of each unit. We further compare shuffled texts with texts produced by probability-constrained generation. 
We adopt word-level shuffling on the natural texts within FOX documents. Taking the Glyph rendering method as an example, the performance achieved by this shuffled-text scheme is relatively higher than that of ZeroSense, which indicates that residual contextual dependencies remain available after shuffling. Probability-constrained generation therefore provides more direct control over contextual predictability than heuristic shuffling. Full experimental details of the shuffling approach are provided in the Appendix.

\begin{table}[t]
\centering
\small
\setlength{\tabcolsep}{12pt}
\begin{tabular}{c ccc}
\toprule
\textbf{Compression} & \multicolumn{3}{c}{\textbf{Threshold}} \\
\cmidrule(lr){2-4}
\textbf{Ratio} & \textbf{$10^{-3}$} & \textbf{$10^{-4}$} & \textbf{$10^{-6}$} \\
\midrule
$2.5\times$  & 0.850 & 0.829 & 0.709 \\
$5.0\times$  & 0.784 & 0.747 & 0.604 \\
$7.5\times$  & 0.707 & 0.659 & 0.453 \\
$10.0\times$ & 0.655 & 0.581 & 0.361 \\
$12.5\times$ & 0.628 & 0.541 & 0.245 \\
$15.0\times$ & 0.588 & 0.491 & 0.246 \\
\bottomrule
\end{tabular}
\caption{Experimental results under varying compression ratios with different probability thresholds (Glyph layout).}
\label{tab:zeroprior_threshold_cr}
\end{table}

We sample low conditional-probability tokens under probability constraints and tune the threshold while keeping the generation model and validation pipeline fixed.

As the threshold decreases from $10^{-3}$ to $10^{-6}$, the OCR accuracy declines consistently. This trend reveals that tighter probability constraints reduce the amount of valid contextual information available for semantic reasoning. Theoretically, we could sweep more threshold values for comprehensive analysis; however, the benchmark relies on cross-validation across multiple evaluation models. It is extremely difficult to find tokens that satisfy a conditional probability threshold stricter than $10^{-6}$ for all models simultaneously. Combined with the heavy computational overhead of dataset generation, we adopt a threshold of $10^{-5}$ as the verification condition for comparative analysis in our experiments.




\section{Conclusion}
In this paper, we investigated the existing evaluation protocols for visual-text compression methods and identified a critical oversight: the coupling of text preservation quality with the inherent linguistic priors of MLLMs. We argued that reliance on downstream task performance leads to biased assessments. To address this, we introduced a novel evaluation framework designed to reduce these confounding factors. The ZeroSense benchmark with low semantic correlation is further introduced to ensure that evaluation results purely reflect visual-text preservation. The extensive experiments reveal a significant divergence between VTC quality and downstream task accuracy. Our work not only offers a clearer understanding of current VTC methods but also establishes a more reliable foundation for the development of future long-context modeling architectures.

\clearpage

\bibliographystyle{unsrt}  
\bibliography{references}

\clearpage

\appendix

\renewcommand{\thetable}{A\arabic{table}}
\setcounter{table}{0}
\renewcommand{\thefigure}{A\arabic{figure}}
\setcounter{figure}{0}

\setcounter{secnumdepth}{0} 
\newcommand{\xyz}[1]{{\color{red} \textbf{\small[XYZ: #1]}}}

\section{A. Implementation Examples of ZeroSense}
\begin{figure*}
    \centering
    \includegraphics[width=0.78\linewidth]{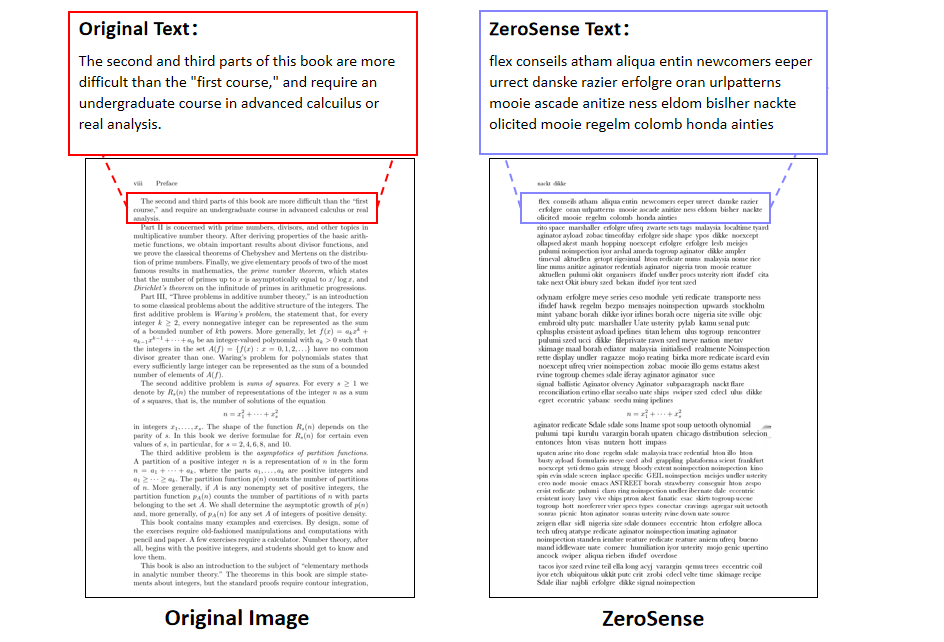}
     \caption{Comparison between original and ZeroSense document images. \textbf{Top:} Full page views demonstrate that our generation pipeline perfectly preserves the document's structural context. \textbf{Bottom:} Transcribed zoomed-in regions highlight the semantic decoupling; true semantic priors are systematically replaced with tokens sampled from a low-probability vocabulary subset, isolating the visual layout characteristics.}
    \label{fig:placeholder}
\end{figure*}

Fig.\ref{fig:placeholder} shows the differences between ZeroSense and natural text datasets. Under the same rendering method, the text in ZeroSense consists of low-semantic-relevance content, making it nearly impossible for models to guess the next token based on contextual semantics. 

\section{B. Full Results under Our Evaluation Framework}

\begin{table*}[t]
\centering
\small
\setlength{\tabcolsep}{5pt}
\begin{tabular}{lccc}
\toprule
\textbf{Setting} &
\textbf{DeepSeek-OCR} &
\textbf{HunyuanOCR} &
\textbf{Qwen2.5-VL-7B} \\
\midrule


Vision encoder &
CLIP-L/14 + SAM-B &
27-layer ViT&
32-layer ViT \\

Visual-token stride &
64 &
32 &
28 \\

Native visual tokens &
256 &
1,058 &
1,369 \\

\bottomrule
\end{tabular}
\caption{Evaluation models and their main visual-token configurations. Native visual-token counts refer to the counts produced under each model's default image-processing configuration.}
\label{tab:model_settings}
\end{table*}

\begin{table*}[t]
\centering
\small
\setlength{\tabcolsep}{5pt}
\begin{tabular}{lccc}
\toprule
\textbf{Setting} &
\textbf{DeepSeek-OCR} &
\textbf{HunyuanOCR} &
\textbf{Qwen2.5-VL-7B} \\
\midrule

Precision &
bfloat16 &
bfloat16 &
bfloat16 \\

\texttt{max\_model\_len} &
4,096 &
4,096 &
8,192 \\

\texttt{max\_num\_seqs} &
256 &
128 &
128 \\

\texttt{gpu\_memory\_utilization} &
0.85 &
0.85 &
0.85 \\

\bottomrule
\end{tabular}
\caption{Inference parameters used for the evaluation models.}
\label{tab:inference_settings}
\end{table*}

\begin{table*}[t]
\centering
\small
\setlength{\tabcolsep}{6pt}
\begin{tabular}{l p{0.72\linewidth}}
\toprule
\textbf{Model} & \textbf{OCR Prompt} \\
\midrule

DeepSeek-OCR &
\texttt{<image>\textbackslash n Free OCR.} \\

HunyuanOCR &
\texttt{OCR the document.} \\

Qwen2.5-VL-7B &
\texttt{Read all the text in this image. Output the text exactly as it appears, line by line. Do not add any explanation.} \\

\bottomrule
\end{tabular}
\caption{OCR prompts used for complete text recovery.}
\label{tab:ocr_prompts}
\end{table*}

\begin{table*}
\centering

\small
\setlength{\tabcolsep}{4pt}
\begin{tabular}{llccccccccc}
\toprule
& &
\multicolumn{3}{c}{DeepSeek-OCR} &
\multicolumn{3}{c}{HunyuanOCR} &
\multicolumn{3}{c}{Qwen2.5VL-7B} \\
\cmidrule(lr){3-5}
\cmidrule(lr){6-8}
\cmidrule(lr){9-11}
Rendering Setting & Compression Ratio
& FOX & Omni & ZeroSense
& FOX & Omni & ZeroSense
& FOX & Omni & ZeroSense \\
\midrule

\multirow{6}{*}{FOX layout}
& \(5\times\)    & 97.7 & 75.2 & \textcolor{red}{33.8} & 66.3 & 40.1 & 2.7 & 81.7 & 44.1 & -- \\
& \(7.5\times\)  & 96.6 & 70.1 & \textcolor{red}{20.2} & 56.2 & 26.7 & -- & -- & 32.5 & -- \\
& \(10\times\)   & 95.4 & 65.3 & \textcolor{red}{11.8} & -- & 23.5 & -- & -- & 35.0 & -- \\
& \(12.5\times\) & 93.8 & 61.2 & \textcolor{red}{7.0} & -- & 19.2 & -- & -- & -- & -- \\
& \(15\times\)   & 91.8 & 56.6 & \textcolor{red}{4.8} & -- & -- & -- & -- & -- & -- \\
& \(17.5\times\) & 87.0 & 51.4 & \textcolor{red}{2.1} & -- & -- & -- & -- & -- & -- \\

\midrule

\multirow{6}{*}{Omni layout}
& \(5\times\)    & 90.7 & 76.7 & 29.1 & 50.4 & 58.9 & 13.3 & 72.9 & 73.9 & \textcolor{red}{42.8} \\
& \(7.5\times\)  & 82.7 & 76.0 & \textcolor{red}{17.1} & 47.2 & 58.8 & 12.0 & -- & 74.2 & -- \\
& \(10\times\)   & 74.1 & 75.6 & \textcolor{red}{12.2} & -- & 60.1 & -- & -- & 78.3 & -- \\
& \(12.5\times\) & 68.6 & 68.6 & \textcolor{red}{11.3} & -- & 59.3 & -- & -- & 78.7 & -- \\
& \(15\times\)   & 61.5 & 61.5 & \textcolor{red}{11.3} & -- & 57.6 & -- & -- & 80.5 & -- \\
& \(17.5\times\) & 57.3 & 57.3 & \textcolor{red}{11.2} & -- & 50.1 & -- & -- & 84.2 & -- \\

\midrule

\multirow{6}{*}{Glyph rendering}
& \(5\times\)    & 98.9 & 94.2 & \textcolor{red}{65.3} & 74.4 & 81.5 & 22.9 & 79.1 & 84.9 & -- \\
& \(7.5\times\)  & 98.2 & 92.3 & \textcolor{red}{52.0} & 58.0 & 68.8 & -- & 93.9 & 76.4 & -- \\
& \(10\times\)   & 97.4 & 90.2 & \textcolor{red}{39.6} & 42.0 & 67.8 & -- & 64.4 & 65.3 & -- \\
& \(12.5\times\) & 96.7 & 87.9 & \textcolor{red}{29.7} & 35.1 & 66.0 & -- & 60.8 & 61.1 & -- \\
& \(15\times\)   & 94.4 & 86.3 & \textcolor{red}{23.5} & 33.5 & 63.0 & -- & -- & 55.3 & -- \\
& \(17.5\times\) & 92.3 & 84.0 & \textcolor{red}{15.1} & -- & 58.0 & -- & -- & 45.8 & -- \\

\bottomrule
\end{tabular}
\caption{
Accuracy across different rendering methods and compression ratios.
FOX and OmniDocBench are natural-text controls, while ZeroSense is used for context preservation evaluation.
For each rendering method, compression ratio, and text source, the highest accuracy among the three models is shown in red. Symbol ``--'' indicates that the rendering method cannot support the corresponding compression ratio due to insufficient model capacity.
}
\label{tab:overall_results2}
\end{table*}
\subsection{B.1 
Model Versions, Inference Parameters, and Prompts}

We evaluate all rendering methods using three MLLMs: DeepSeek-OCR, HunyuanOCR, and Qwen2.5-VL-7B. No model is fine-tuned or adapted on ZeroSense. Unless otherwise specified, inference is conducted with vLLM 0.23.0 and PyTorch 2.11.0 in bfloat16 precision on NVIDIA RTX A6000 GPUs with 48 GB memory. The model configurations are summarized in Table~\ref{tab:model_settings}, the inference parameters are reported in Table~\ref{tab:inference_settings}, and the OCR prompts are listed in Table~\ref{tab:ocr_prompts}.

\subsection{B.2 
Detailed Numerical Context Preservation of Different Methods}
Tab.\ref{tab:overall_results2} presents the results across different rendering methods on ZeroSense.
In addition, natural text from Fox and Omni was used as a baseline for comparison. 
The best contextual preservation achieved at each compression ratio across all evaluated models is highlighted in red.

It can be observed that the performance of all VTC methods drops significantly when evaluated on the ZeroSense dataset. 
Even with Fox layout rendering method at the relatively low compression ratio of 5×,
the OCR retention rate drops sharply from 97.7\% on natural text to 33.8\%, and further collapses to 2.1\% at a 17.5× compression ratio. This underscores the necessity of employing ZeroSense when evaluating contextual information preservation.

Due to its architecture, Qwen2.5-VL produces substantially more visual tokens during image encoding, making higher compression ratios infeasible with a fixed input text length. As a result, many entries corresponding to higher compression ratios are unavailable in the Qwen2.5-VL column of Table \ref{tab:overall_results2}.
However, it is worth noting that Qwen2.5-VL exhibits stronger performance at the 5× compression ratio with Omni layout rendering method. 
It suggests that certain models are inherently better suited to specific rendering methods. 
Consequently, the context preservation capability of a rendering method cannot be reliably assessed using only a single evaluation model.
This demonstrates the necessity of multi-model evaluation in our framework. 

\subsection{B.3 Ablation Study on Effect of Context Length}
\begin{table*}
\centering

\begin{tabular}{lccccc}
\toprule
 & & \multicolumn{4}{c}{\textbf{Text Length}} \\
\cmidrule(lr){3-6}
\textbf{Rendering Setting} & \textbf{Compression Ratio} & \textbf{200--400} & \textbf{400--600} & \textbf{600--800} & \textbf{800--1000} \\
\midrule
\multirow{6}{*}{FOX layout} 
 & 2.5$\times$  & 46.2 & 44.2 & 34.3 & 28.5 \\
 & 5$\times$  & 28.3 & 38.5 & 35.0 & 28.5 \\
 & 7.5$\times$  & 17.4 & 21.4 & 20.8 & 21.2 \\
 & 10$\times$ & 13.7 & 11.7 & 11.3 & 6.5 \\
 & 12.5$\times$ & --   & 7.9  & 6.3  & 4.3 \\
 & 15$\times$ & --   & 6.2  & 4.9  & --  \\

\midrule
\multirow{6}{*}{Omni layout} 
 & 2.5$\times$  & 48.5 & 38.7 & 37.3 & 40.1 \\
 & 5$\times$  & 26.9 & 29.7 & 30.9 & 28.5 \\
 & 7.5$\times$  & 16.1 & 15.6 & 18.8 & 21.1 \\
 & 10$\times$ & 12.9 & 10.4 & 12.8 & 17.1 \\
 & 12.5$\times$ & 11.8 & 11.0 & 10.4 & 14.0 \\
 & 15$\times$ & --   & 9.7  & 10.5 & 13.5 \\
 
\midrule
\multirow{6}{*}{Glyph rendering} 
 & 2.5$\times$  & 71.7 & 68.3 & 64.4 & 59.1 \\
 & 5$\times$  & 67.8 & 65.8 & 63.6 & 57.6 \\
 & 7.5$\times$  & 54.8 & 51.8 & 50.0 & 48.2 \\
 & 10$\times$ & 42.7 & 39.2 & 37.6 & 36.2 \\
 & 12.5$\times$ & 32.2 & 29.1 & 27.3 & 27.4 \\
 & 15$\times$ & 27.1 & 25.4 & 23.4 & 21.1 \\
 
\bottomrule
\end{tabular}
\caption{Maximum accuracy across different compression ratios and text lengths on ZeroSense.}
\label{tab:ablation_text_length}
\end{table*}
Table \ref{tab:ablation_text_length} reports the sensitivity of different rendering methods to the input text length across different compression ratios.
The results indicate that the performance of different rendering methods varies much with input text length.
Specifically, both Fox and Glyph rendering methods 
show a consistent decline in OCR performance
as text length increases from 200 to 1,000 tokens. 
In other words,
under a fixed compression ratio, a longer text input always leads to lower OCR performance. In contrast, Omni layout 
does not exhibit a clear monotonic trend, and in some cases even achieves better performance for input lengths between 800 and 1,000 tokens.
Therefore, further investigation into the interaction between rendering methods and input text length is warranted.

\subsection{B.4 Comparison with Heuristic Text Perturbations}
\label{sec:word_perturbation}

To examine whether simple character-level perturbations can provide a similarly evaluation benchmark or server as simpler alternatives to the proposed ZeroSense, we additionally construct two heuristic baselines,
Swap and Shuffle, by modifying natural English text from FOX.

\paragraph{Perturbation Strategies.}
Both strategies first select a specified proportion of eligible words from the source text. A word is considered eligible if it contains more than one alphabetic character
after removing surrounding punctuation. Word positions and surrounding punctuation are kept unchanged.

For \emph{Swap}, two randomly selected characters within each chosen word are exchanged. This operation preserves the character set and most of the original word structure while introducing a mild spelling disturbance. For \emph{Shuffle}, all characters within each chosen word are randomly permuted. This operation preserves the character set and word length but more strongly disrupts within-word character order. Table~\ref{tab:swap_shuffle_examples} illustrates the difference between the two strategies.

\begin{table*}[t]
\centering
\small
\setlength{\tabcolsep}{6pt}
\renewcommand{\arraystretch}{1.15}
\begin{tabular}{p{0.28\linewidth} p{0.28\linewidth} p{0.28\linewidth}}
\toprule
\textbf{Original Text} &
\textbf{Swap Perturbation} &
\textbf{Shuffle Perturbation} \\
\midrule

The committee unanimously approved the annual budget proposal after a brief discussion.
&
The committee unanimously \underline{avproped} the annual
\underline{gudbet} proposal after a brief discussion.
&
The committee unanimously \underline{prvapoed} the annual
\underline{uetdbg} proposal after a brief discussion.
\\

\bottomrule
\end{tabular}
\caption{Examples of the Swap and Shuffle perturbation strategies. Swap exchanges two randomly selected characters within each chosen word, whereas Shuffle randomly permutes all characters within the chosen word. Modified words are underlined.}
\label{tab:swap_shuffle_examples}
\end{table*}

For each strategy, we perturb \(p\in\{5,10,50,100\}\%\) of the eligible words, resulting in eight settings: Swap5, Swap10, Swap50, Swap100, Shuffle5, Shuffle10, Shuffle50, and Shuffle100. All random operations use a fixed seed of 0 for reproducibility.

\paragraph{Experimental Setup.}
We apply the eight perturbation settings to the ground-truth English texts of 112 FOX documents. The resulting texts contain approximately 650--3,500 tokens per document under the DeepSeek tokenizer. Each perturbed text is rendered using the Glyph rendering pipeline with the same configuration as in the main experiments. We use only Glyph rendering in this experiment to keep the layout fixed across all text conditions. Each perturbation setting produces 112 rendered document images.

The rendered images are evaluated using DeepSeek-OCR, HunyuanOCR, and Qwen2.5-VL-7B. We use the same inference settings and OCR prompts as those reported in Section~B.1. 
Following the main evaluation protocol, we compute the word-level set precision for each model and report the maximum 
score across the three evaluation models.

\begin{table*}[t]
\centering
\small
\setlength{\tabcolsep}{7pt}
\begin{tabular}{lcccccc}
\toprule
\textbf{Text Condition}
& \(\mathbf{5\times}\)
& \(\mathbf{7.5\times}\)
& \(\mathbf{10\times}\)
& \(\mathbf{12.5\times}\)
& \(\mathbf{15\times}\)
& \(\mathbf{17.5\times}\) \\
\midrule

FOX text (clean)
& 98.9 & 98.2 & 97.4 & 96.7 & 94.4 & 92.3 \\

\midrule

Swap5
& 84.9 & 82.4 & 77.2 & 73.0 & 71.0 & 73.0 \\

Swap10
& 81.9 & 80.6 & 74.2 & 74.4 & 66.7 & 71.6 \\

Swap50
& 67.2 & 64.5 & 62.2 & 57.7 & 57.9 & 51.8 \\

Swap100
& 57.2 & 39.9 & 49.7 & 42.4 & 51.1 & 29.6 \\

\midrule

Shuffle5
& 84.2 & 82.7 & 76.3 & 73.4 & 70.3 & 70.4 \\

Shuffle10
& 81.2 & 79.8 & 75.0 & 75.0 & 64.6 & 71.2 \\

Shuffle50
& 65.5 & 62.3 & 65.0 & 63.8 & 53.5 & 57.6 \\

Shuffle100
& 58.4 & 46.1 & 45.3 & 58.0 & 45.7 & 51.3 \\

\bottomrule
\end{tabular}
\caption{Maximum accuracy across DeepSeek-OCR, HunyuanOCR, and Qwen2.5-VL-7B for clean (original texts without any scrambling) and heuristically perturbed FOX text under Glyph rendering. Swap\(p\) and Shuffle\(p\) perturb \(p\%\) of eligible words. All conditions use the same rendering method, compression ratios, evaluation models, and aggregation protocol.}
\label{tab:perturbation_results}
\end{table*}
\paragraph{Results.}
Table~\ref{tab:perturbation_results} shows that both Swap and Shuffle
substantially reduce OCR 
performance.
 Increasing the perturbation ratio generally leads to lower recovery
scores, indicating that the linguistic and orthographic regularities
of natural text contribute to the high OCR accuracy.

We further compare these heuristic perturbations with the ZeroSense
results under Glyph rendering reported in
Table~\ref{tab:overall_results2}. At \(10\times\) compression, Swap100
and Shuffle100 achieve 0.497 and 0.453, respectively, compared with
0.396 for ZeroSense. At \(15\times\), the corresponding scores are
0.511 and 0.457, whereas ZeroSense achieves 0.235. These results indicate
that even strong within-word perturbations preserve more recoverable linguistic
structure than ZeroSense under medium and high compression.

Swap and Shuffle preserve the original word order, sentence structure,
punctuation, word lengths, and character composition, 
leaving substantial contextual information and partial orthographic cues that can still facilitate recognition.
In contrast, ZeroSense reduces predictability throughout the source text
through probability-constrained generation and cross-model validation.
The results therefore indicate that heuristic perturbations weaken lexical regularities but do not remove the broader contextual information available to OCR models, whereas ZeroSense suppresses predictability throughout the document.

\end{document}